\documentclass{article}
\usepackage{spconf,amsmath,graphicx}

\usepackage{times}
\usepackage{epsfig}
\usepackage{amsmath}
\usepackage{amssymb}
\usepackage{multirow}
\usepackage{subcaption}
\usepackage{array}
\usepackage{xcolor}
\usepackage{ifthen}
\usepackage{adjustbox}
\usepackage{enumitem}
\usepackage[pagebackref=true,breaklinks=true,colorlinks,bookmarks=false]{hyperref}

\newcolumntype{!}{>{\global\let\currentrowstyle\relax}}

\newcolumntype{^}{>{\currentrowstyle}}
\newcommand{\rowstyle}[1]{\gdef\currentrowstyle{#1}%

    #1\ignorespaces

}

\providecommand{\red}[1]{\textcolor{red}{#1}}

\def\debug{1}

\ifnum \debug=1
\providecommand{\lnote}[1]{\textcolor{blue}{[LR: #1]}}
\providecommand{\tnote}[1]{\textcolor{teal}{[TF: #1]}}
\providecommand{\vnote}[1]{\textcolor{orange}{[VB: #1]}}
\providecommand{\anote}[1]{\textcolor{red}{[AP: #1]}}
\providecommand{\mnote}[1]{\textcolor{violet}{[MB: #1]}}
\else
\providecommand{\lnote}[1]{}
\providecommand{\tnote}[1]{}
\providecommand{\vnote}[1]{}
\providecommand{\anote}[1]{}
\providecommand{\mnote}[1]{}
\fi

\title{Memory-augmented Online Video Anomaly Detection}
\name{Leonardo Rossi \qquad Vittorio Bernuzzi \qquad Tomaso Fontanini \qquad Massimo Bertozzi \qquad Andrea Prati}
\address{University of Parma, Department of Engineering and Architecture}

\begin{document}
%
\maketitle
%
    \title{Memory-augmented Online Video Anomaly Detection}
    \maketitle
    %

    \begin{abstract}
        The ability to understand the surrounding scene is of paramount importance for Autonomous Vehicles (AVs).
This paper presents a system capable to work in an online fashion, giving an immediate response to the arise of anomalies surrounding the AV, exploiting only the videos captured by a dash-mounted camera.
Our architecture, called \emph{MOVAD}, relies on two main modules: a Short-Term Memory Module to extract information related to the ongoing action, implemented by a Video Swin Transformer (VST), and a Long-Term Memory Module injected inside the classifier that considers also remote past information and action context thanks to the use of a Long-Short Term Memory (LSTM) network. 
The strengths of MOVAD are not only linked to its excellent performance, but also to its straightforward and modular architecture, trained in a end-to-end fashion with only RGB frames with as less assumptions as possible, which makes it easy to implement and play with.
We evaluated the performance of our method on Detection of Traffic Anomaly (DoTA) dataset, a challenging collection of dash-mounted camera videos of accidents.
After an extensive ablation study, MOVAD is able to reach an AUC score of 82.17\%, surpassing the current state-of-the-art by $+2.87$ AUC.
    \end{abstract}

    \section{Introduction and related works}

Autonomous Vehicles (AVs) are becoming every day a reality thanks to the recent enormous scientific and technical advances.
Nevertheless, safety of AVs is still a relevant issue which can jeopardize their world-wide diffusion.
Increasing safety of AVs can be reached by providing vehicles with the ability of detecting anomalous situations in a prompt way.
Their detection provides information to avoid collisions, protect pedestrians, or re-route the current travel \cite{4298901}.
Among the different sensors exploited in AVs, cameras offer rich real-time information about the scene, but, despite all, anomalous situation detection in real traffic scenarios is still very challenging.
In particular, we still miss a formal shared model of what an anomaly should be, because many times the riskiness is highly subjective.
In addition, there are plenty of possible accident classes with a very exiguous number of examples to take into account compared to normal traffic situations.
Lastly, the definition of time boundaries of an anomaly is even more subjective and doubtful.
Nevertheless, some attempts have been made to propose a deterministic method to define an anomaly.
Fang \emph{et al.} \cite{fang2019dada} labels the anomaly start from the moment in which half part of the object involved in the accident appears in the view.
Yao \emph{et al.}~\cite{9712446} proposed the Detection of Traffic Anomaly (DoTA) dataset, 
where the anomaly is starting the instant after which the accident is unavoidable. 
We take this last dataset as the benchmark for our work because it is the most complete and used dataset for this type of task in the road traffic safety context.

There have been several previous works addressing the problem of video anomaly detection.
Authors in \cite{hasan2016learning} proposed a Convolutional AutoEncoder (ConvAE) trained only on normal frames with the objective of frame reconstruction.
In \cite{luo2017remembering,wang2018abnormal}, authors used Convolutional LSTM Auto-Encoder as framework to encode appearance and motion.
As noted by \cite{ramachandra2020survey}, auto-encoder-type reconstruction methods are sensitive to the amount of anomalies that occurs in the scene and many times they require additional ad-hoc post-processing techniques.
Authors in \cite{liu2018future} proposed AnoPred, which uses a multi-task loss, including image intensity, optical flow, gradient and adversarial losses for Video Anomaly Detection (VAD) by predicting a future frame.
As stated in~\cite{9712446}, AnoPred was thought in a video surveillance context, while videos acquired from inside a moving vehicle are more dynamic and difficult to predict.
In \cite{zhou_spatio-temporal_2022}, authors of STFE model make a two-stage detector: a coarse detection to encode temporal features with Histogram of Optical Flow (HOF) \cite{wang2013action} and ordinal features of frames by a CNN, and then, a fine detection, by encoding the CNN features and spatial relationships of the objects.
In FOL~\cite{9712446}, authors try to avoid the future prediction for the entire frame, focusing instead on tracking actors' positions and predicting their future locations.
Conversely to us, the applicability of both last methods are limited by the presence of actors in the Field of View (FOV).
Moreover, AnoPred~\cite{liu2018future}, STFE \cite{zhou_spatio-temporal_2022} and FOL \cite{9712446} rely on supplementary input information other than RGB frames, such as optical flow, bounding boxes and ego-motion information of the moving camera.
Our model outperforms them by exclusively processing RGB frames, thereby highlighting its capability to comprehend anomalies in video traffic scenes without the need for auxiliary information.
In~\cite{xu2019temporal}, the TRN model couples the action detection task with both the temporal dependencies modeled by a RNN and the anticipation of the future via a temporal decoder.
Like them, we exploit an RNN to obtain temporal information, but not limiting ourselves to this.
We formulated an advanced version of memory, differentiating between short- and long-memory through two distinct and trained-at-same-time modules, obtaining a more informative description of the ongoing situation, without speculating on the future.
Summarizing, the main contributions of this paper are:
\begin{itemize}
    \item A straightforward end-to-end architecture for Online Video Anomaly Detection (VAD) called \emph{MOVAD}, which processes ongoing RGB frames, with as less assumptions as possible about the ongoing action and without any extra auxiliary information.
    \item A plugged-in Transformer ~\cite{liu_video_2022} model used as backbone for Short-Term Memory information extraction in an online scenario.
    This allows the system to incorporate the most recent temporal and spatial correlations by relying only on the current and few past frames.
    \item The injection of a LSTM module inside the classification head to model the Long-Term Memory and to exploit contextual information spread across the entire history.
    \item An exhaustive ablation study over the dataset DoTA~\cite{9712446}.
    Compared w.r.t.~the state of the art, MOVAD shows superior performance in terms of AUC.
\end{itemize}
    \section{Memory-augmented Online VAD}
\label{sec:theory}

\fboxsep=1mm
\fboxrule=1pt


\begin{figure}[!t]
            \centerline{\includegraphics[clip, trim=236 510 113 133, width=\linewidth]{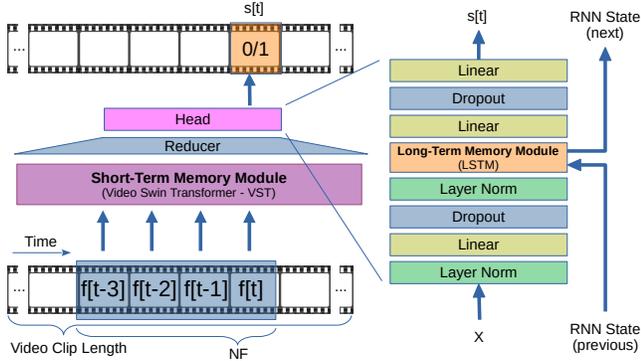}}
        \caption{The online frame-level VAD architecture. $f[t]$ is the frame at time $t$, $x$ the output of the Reducer, $\mathit{NF}$ the number of frames in input to the VST, $s[t]$ the anomaly classification score of the frame $f[t]$.\label{fig:arch}}
\end{figure}

In our system, recently observed frames and past frames are taken into account as a source of information related to the ongoing action and the context, respectively.
In order to process current and past frames for the Online VAD task, we equipped MOVAD of two main components: a Short-Term Memory Module (STMM) and a Long-Term Memory Module (LTMM) inside the classification head (see Fig.~\ref{fig:arch}). 

\noindent\textbf{Short-Term Memory Module.}
In order to pre-elaborate the new coming frame $f[t]$ and, in conjunction, to retain near-past spatio-temporal information $f[t-1] .. f[t-\left(\mathit{NF}-1\right)]$, we chosen a transformer architecture over an RNN, to be able to process them in a parallel fashion.
We lean towards transformer over 3D convolutional models because of the well demonstrated abilities to intercept long-distance interactions in space and time \cite{moutik2023convolutional}.
We selected VST~\cite{liu_video_2022} as our backbone network for the STMM over models like ViViT~\cite{Arnab_2021_ICCV}, due to its superior performance and more efficient computation of self-attention mechanism.
Despite this, the self-attention mechanism remains computationally intensive, particularly for long videos. 
Therefore, we limited the input to a small temporal window of $\mathit{NF} = 4$ frames of the video, going from the current frame at time $t$ to the previous frame at time $t-\left(\mathit{NF}-1\right)$.
In this way, we obtain a compact representation of the short history that will be subsequently forwarded to the LTMM.
VST takes as input a video with size $\mathit{NF} \times H \times W \times 3$, where $\mathit{NF}$, $H$, $W$ and $3$ correspond to the number of frames, height, width and RGB channels, respectively.
The model internally splits the frames in non-overlapping 3D patches, partitioning the video in $\frac{\mathit{NF}}{2} \times \frac{H}{4} \times \frac{W}{4}$ 3D tokens, and using a 3D shifted window mechanism to obtain cross-window connections and exploit spatio-temporal information.

\noindent\textbf{Long-Term Memory Module.}
\label{Long-Term Memory Module descr}
As will be highlighted in ablation in Section~\ref{Short-Term Memory Module exp}, too much information about the past ($\mathit{NF} > 4$) tends to mislead the STMM, resulting in a loss of performance.
According to our hypothesis, this occurs because all frames are placed at the same level, without a possibility of weighing them based on how much they belong to the past.
For this reason, we engineered a different way to integrate and enrich the output latent space from the STMM with contextual information extrapolated from the distant past.
Every time a new frame is available, we transfer the compact representation obtained from the STMM to our LTMM, in order to update its representation of the past.
In this way, we can maintain the focus on what is happening now, and at same time accumulating a richer understanding of the scene, enabling more precise classifications.
In details, the output of VST goes through Adaptive Average Pool 3D layer (Reducer in Fig.~\ref{fig:arch}) and, finally, enters inside the classification head.
As shown in Fig.~\ref{fig:arch}, the head is composed by a series of normalization layers, linear layers and dropout, alternating. 
A stacked three-cell LSTM module is inserted after the last normalization layer to model the long-term memory.
The state, composed by three hidden $h[t]$ and cell $c[t]$ states, is updated whenever a new frame is available.
The LSTM receives in input a features block of $[B, 1024]$, where $B$ is the batch size, and returns a block of same size together with the state of the cells.
Since the state is relatively small, the module is very efficient and leads to a fixed and limited additional computational cost.
For each frame $f[t]$, the classification head outputs the anomaly classification score $s[t] \in [0,1]$, where $0$ means no anomaly and $1$ means the frame is anomalous.
A weighted cross-entropy loss was chosen to train the model, giving higher weight to the anomaly class reflecting the distribution of the data.
We used the formula $w_i = e / e_i$, where $w_i$ is the weight associated to the class, $e$ the total amount of examples in the dataset and $e_i$ the amount of examples for the class $i$.

    \section{Experimental Results}
\label{sec:experiments}

\noindent\textbf{Dataset}.
We perform our training and test on Task~1, the frame-level VAD, of DoTA dataset \cite{9712446}, using only the anomaly class and its temporal boundaries, strictly in the online scenario.

\newcommand{\figsize}{0.8\columnwidth}

\begin{figure}[t!]
\centerline{\includegraphics[clip,width=\figsize]{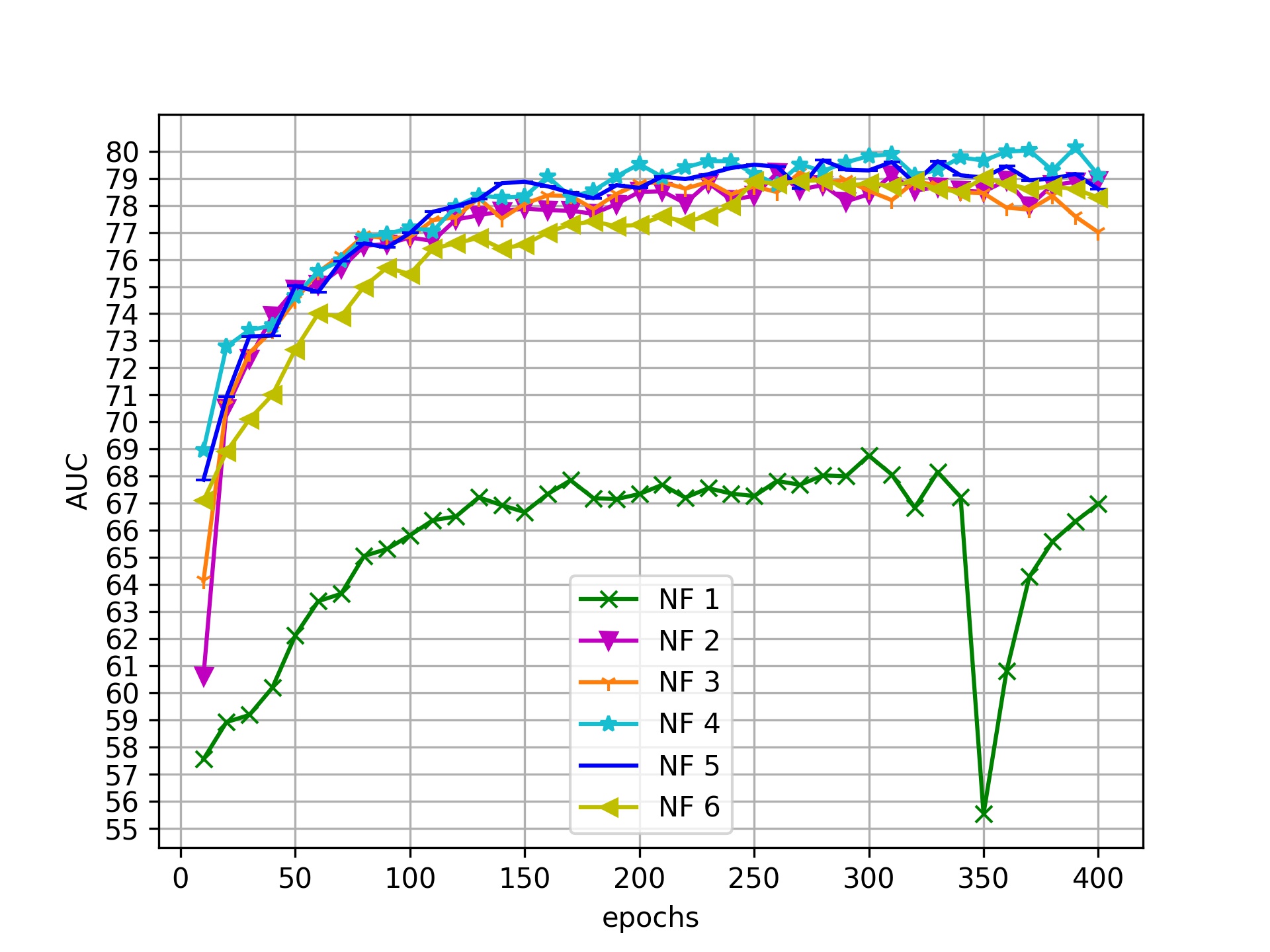}}
	\caption{Performance comparison, changing the number of frames ($\mathit{NF}$) in input to the STMM (from 1 to 6).}
	\label{fig:num-frames-vst}
\end{figure}

\begin{figure}[t!]
\centerline{\includegraphics[clip,width=\figsize]{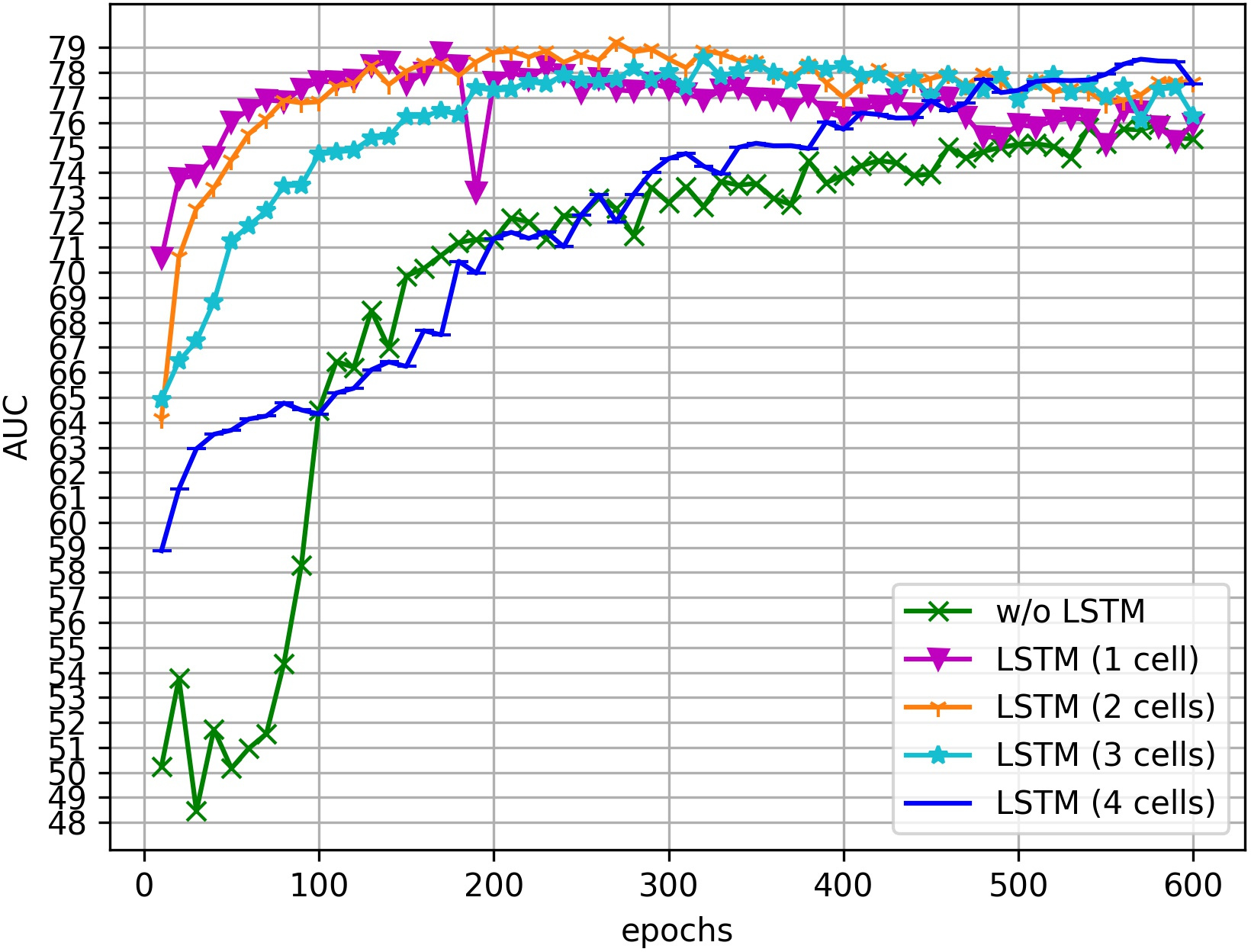}}
	\caption{Performance comparison, changing the number of LSTM cells (from 0 to 4). ``LSTM (2 cells)'' corresponds to ``$\mathit{NF}=3$''  in Fig.~\ref{fig:num-frames-vst}.}
	\label{fig:num-memory-cells}
\end{figure}

\noindent\textbf{Evaluation Metrics}.
We use the well-known Area Under the Curve (AUC) metric at frame-level, to evaluate how well the model is able to temporally locate the anomaly in the videos.

\noindent\textbf{Training details.}
We perform the training on a single machine with an A100 GPU\@, with Stochastic Gradient Descent (SGD) optimization algorithm, learning rate of 0.0001, momentum of 0.9.
We use SGD instead of Adam because in our experiments the latter led the training to be more unstable, resulting in model diverging after few epochs.
Unless otherwise specified, batch size is 8, input video size is $320 \times 240$, Video Clip Length (VCL), which is the number of frames inside the batch for each video, is 8, LSTM cell number is 2, $\mathit{NF}$ is 3, linear weights are initialized using a uniform distribution, LSTM cells with a (semi-)orthogonal matrix, bias parameters are set to zero and the VST is initialized with a model pretrained on Something-Something v2~\cite{goyal2017something}.
We train using a weighted Cross-Entropy loss to address the issue of imbalanced data within the DoTA dataset, assigning $w_n=0.3$ and $w_a=0.7$ to the normal and anomaly class, respectively following equation reported at the end of Section \ref{sec:theory}.

\begin{figure}[ht!]
\centerline{\includegraphics[clip,width=\figsize]{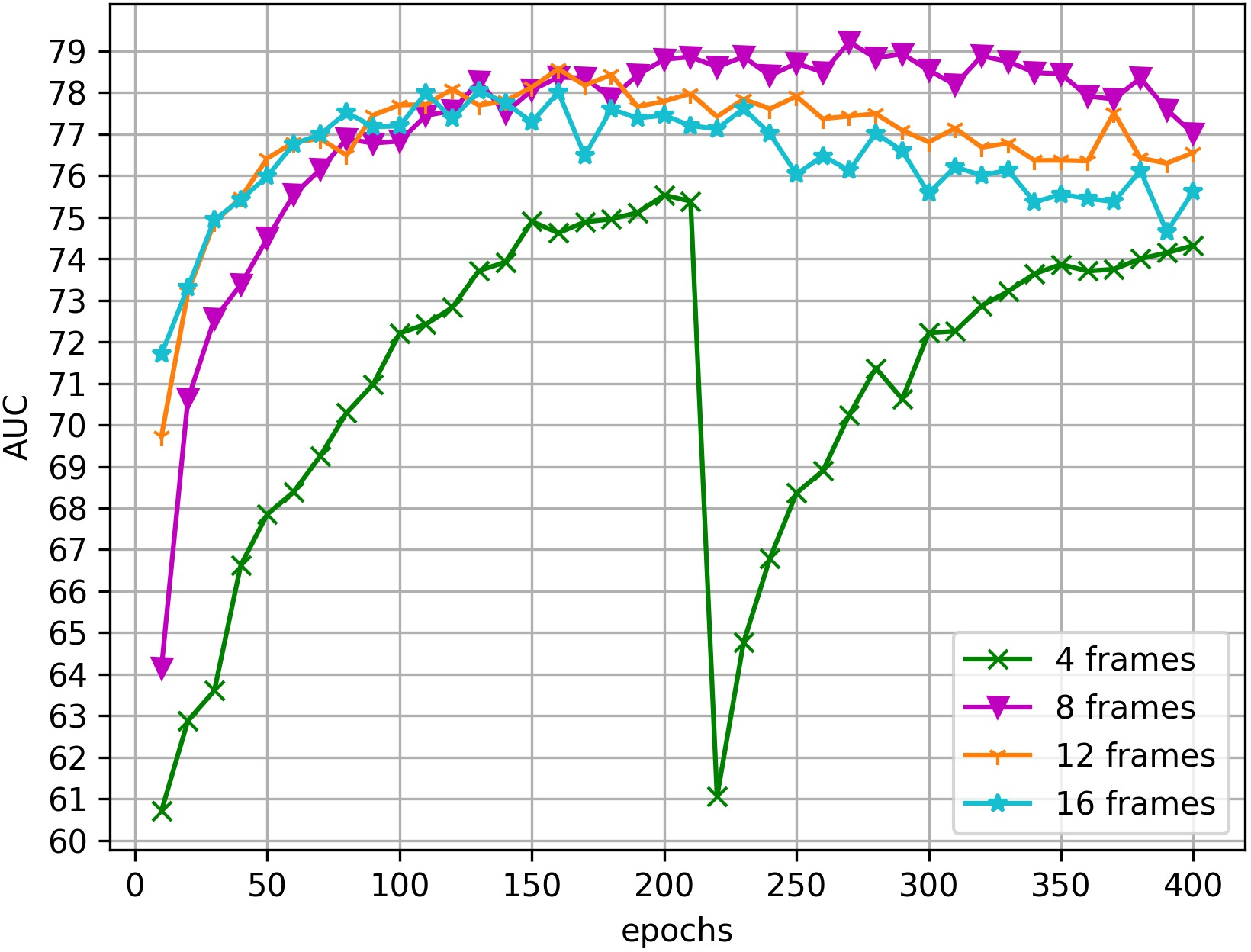}}
    \caption{Performance comparison, changing the VCL. The ``8 frames'' configuration corresponds to ``$\mathit{NF}=3$'' in Fig.~\ref{fig:num-frames-vst}.\label{fig:random-batch}}
\end{figure}
\begin{table}[ht!]
	\normalsize
	\centering
     \begin{adjustbox}{width=0.5\linewidth}
     \begin{tabular}{!r|^c|^c|^c|}
                \# & Short-term & Long-term & $AUC$ \\
                \hline\hline
                        1 &            &            & 66.53 \\
                        2 & \checkmark &            & 74.46 \\
                        3 &            & \checkmark & 68.76 \\
    \rowstyle{\bfseries}    4 & \checkmark & \checkmark & 79.21 \\
    \end{tabular}
    \end{adjustbox}
    \hspace{0.5em}
	\caption{Performance comparison with and w/out short and long-term memory. Short-term: with ($\mathit{NF}=3$) and w/out ($\mathit{NF}=1$). Long-term: with (2 cells) and w/out (0 cells). }
	\label{tab:short-vs-long-term-auc}
\end{table}
\begin{table}[ht!]
	\normalsize
	\setlength{\tabcolsep}{4.2pt}
    
	\centerline{
 \begin{adjustbox}{width=0.85\linewidth}
 \begin{tabular}{!r|^l|^l|^c|}
			\# & Method & Input & $AUC$ \\
			\hline\hline
	   	            1 & ConvAE \cite{hasan2016learning} (*)       & Gray                   & 64.3 \\
			        2 & ConvAE \cite{hasan2016learning} (*)       & Flow                   & 66.3 \\
                    3 & ConvLSTMAE \cite{chong2017abnormal} (*)   & Gray                   & 53.8 \\
                    4 & ConvLSTMAE \cite{chong2017abnormal} (*)   & Flow                   & 62.5 \\
                    5 & AnoPred \cite{liu2018future} (*)          & RGB                    & 67.5 \\
                    6 & AnoPred \cite{liu2018future} (*)          & Masked RGB             & 64.8 \\
                    7 & FOL-Ensemble \cite{9712446}  (*)          & RGB + Box + Flow + Ego & 73.0 \\
                    8 & STFE \cite{zhou_spatio-temporal_2022}     & RGB + Flow             & 79.3 \\
            \hline
\rowstyle{\bfseries}9 & Our (MOVAD)                           & RGB ($320\times240$)   & 80.09 \\
\rowstyle{\bfseries}10 & Our (MOVAD)                          & RGB ($640\times480$)   & 82.17 \\
\end{tabular}
\end{adjustbox}}
    \hspace{0.5em}
	\caption{Benchmarks of VAD methods on the DoTA dataset. Both MOVAD models are trained with the best configuration of: VCL of 8, 3 LSTM cells and $\mathit{NF}=4$. (Note: Results with (*) are taken from DoTA \cite{liu2018future} paper.}
	\label{tab:sota-vad-auc}
\end{table}
\begin{table}[ht!]
	\normalsize
	\setlength{\tabcolsep}{3.5pt}
    \begin{adjustbox}{max width=\linewidth}
	\begin{tabular}{
            |!l||^c|^c|^c|^c|^c|^c|^c|^c|^c|}
            \hline
			Model & $ST$ & $AH$ & $LA$ & $OC$ & $TC$ & $VP$ & $VO$ & $OO$ & $UK$ \\
			\hline\hline
                AnoPred \cite{liu2018future}          & 69.9 & 73.6 & 75.2 & 69.7 & 73.5 & 66.3 & N/A & N/A & N/A  \\
                AnoPred \cite{liu2018future} + Mask   & 66.3 & 72.2 & 64.2 & 65.4 & 65.6 & 66.6 & N/A & N/A & N/A \\
                FOL-STD \cite{9712446}                & 67.3 & 77.4 & 71.1 & 68.6 & 69.2 & 65.1 & N/A & N/A & N/A \\
                FOL-Ensemble \cite{9712446}           & 73.3 & 81.2 & 74.0 & 73.4 & 75.1 & 70.1 & N/A & N/A & N/A \\
                STFE \cite{zhou_spatio-temporal_2022} & 75.2 & 84.5 & 72.1 & 77.3 & 72.8 & 71.9 & N/A & N/A & N/A \\
                \textbf{Our (MOVAD)   }                &  \red{85.6} & \red{85.1} & \red{83.9} & \red{82.2} & \red{85.3} & \textbf{86.2} & \textbf{79.3} & \red{86.7} & \textbf{77.1} \\
                \textbf{Our (MOVAD) \dag}                &  \textbf{86.6} & \textbf{86.3} & \textbf{84.9} & \textbf{83.7} & \textbf{85.5} & \red{81.6} & \red{77.4} & \textbf{87.9} & \red{73.8} \\
                \hline
            \hline
			Model & $ST*$ & $AH*$ & $LA*$ & $OC*$ & $TC*$ & $VP*$ & $VO*$ & $OO*$ & $UK*$ \\
			\hline\hline
                AnoPred \cite{liu2018future}          & 70.9 & 62.6 & 60.1 & 65.6 & 65.4 & 64.9 & 64.2 & 57.8 & N/A \\
                AnoPred \cite{liu2018future} + Mask   & 72.9 & 63.7 & 60.6 & 66.9 & 65.7 & 64.0 & 58.8 & 59.9 & N/A \\
                FOL-STD \cite{9712446}                & 75.1 & 66.2 & 66.8 & 74.1 & 72.0 & 69.7 & 63.8 & 69.2 & N/A \\
                FOL-Ensemble \cite{9712446}           & \red{77.5} & 69.8 & 68.1 & \red{76.7} & 73.9 & 71.2 & 65.2 & 69.6 & N/A \\
                STFE \cite{zhou_spatio-temporal_2022} & \textbf{80.6} & 65.6 & 69.9 & 76.5 & 74.2 & N/A & 75.6 & 70.5 & N/A \\
                \textbf{Our (MOVAD)   }                & 72.1 & \red{71.6} & \red{72.3} & 76.5 & \red{75.7} & \red{74.1} & \red{77.9} & \red{71.7} & \red{69.1} \\
                \textbf{Our (MOVAD) \dag}              & 72.2 & \textbf{74.0} & \textbf{74.8} & \textbf{80.2} & \textbf{79.6} & \textbf{76.8} & \textbf{82.2} & \textbf{78.3} & \textbf{72.9} \\
        \hline
\end{tabular}
    \end{adjustbox}
    \hspace{0.5em}
	\caption{Detection accuracy (AUC) for individual accident categories. ``*'' non-ego anomaly categories. ``\dag'' if input resolution is $640\times480$ instead of $320\times240$. N/A=Not Available. Bold and red values are the best and second-best results.}
	\label{tab:sota-vad-auc-per-class}
\end{table}
\subsection{Ablation study}
In this section different configurations will be individually explored and, finally, an optimal setup of MOVAD will be presented in Section \ref{comparison-with-sota}.

\noindent\textbf{Memory modules effectiveness.}
As shown in Table~\ref{tab:short-vs-long-term-auc}, we first tested the effect of memories.
STMM and LTMM both contribute to enhance the general performance, obtaining the best AUC when both are active, highlighting their importance. 

\noindent\textbf{Short-Term Memory Module.}
\label{Short-Term Memory Module exp}
We decided to design our STMM based on a Video Swin-B architecture, characterized by an embedding dimension of $C = 128$ after the linear projection of the patches. 
In Fig.~\ref{fig:num-frames-vst}, results when varying the number of frames $\mathit{NF}$ processed by the STMM at each step are displayed.
As expected, taking into account only the current frame is the worst situation, loosing any temporal information and making the training unstable.
This is reasonable: by not having any knowledge of the near past, the STMM leads the LTMM to overfitting since consecutive frames are very similar.
On the contrary, increasing the number of frames generally increase the performance. 
This is true until processing 5 or more frames, where the effect becomes counterproductive.
As mentioned in Section \ref{Long-Term Memory Module descr}, according to our hypotheses, overloading the transformer becomes harmful, as it does not have a mechanism to weight the remote and recent past differently.
Overall, highest AUC is obtained with 4 frames.

\noindent\textbf{Long-Term Memory Module.}
In Fig.~\ref{fig:num-memory-cells} the LTMM capabilities are evaluated, varying the number of cells from zero (no LSTM at all) to four.
Indeed, having no cells makes training slower in saturating performance and the lowest AUC is reached.
In general, by increasing the number of cells, the maximum AUC is reached slower but it is higher than w/out LSTM.
With 1 cell the performance saturates very quickly, with a slow degradation during the rest of the epochs.
On the other side, 4 cells lead to a very slow saturation w/out reaching the best performance.
The global maximum AUC is reached with 2 cells, even if with just +$0.02$ compared to 3 cells.
Despite this, we prefer the latter configuration, because we think its ability to increase the quality of training in a slower and more continuous way has a better general benefit.
We verified this hypothesis and, in conjunction with $\mathit{NF}=4$ (best configuration in the previous experiment), it permits to obtain higher AUC than with 2 cells.
We speculate this happens because both (3 cells in Fig. \ref{fig:num-memory-cells} and $\mathit{NF}=4$ in Fig. \ref{fig:num-frames-vst}) reach the best AUC at same time (around 400 epochs).


\noindent\textbf{Video clip length (VCL).}
In Fig.~\ref{fig:random-batch}, different values of VCL are explored.
The worst and most unstable training is obtained with 4 frames, probably because they are too few to exploit the long-term memory effect of LSTM cells.
Increasing VCL permits to saturate performance quicker, but a value too high (like 12 or 16) tends to produce a lower AUC overall.
The highest AUC is obtained with 8 frames as VCL, which represents a good trade-off between enlarging clip size and exploiting LSTM cells, and reducing it to avoid overfitting, due to consecutive frames being too similar to each other.

\subsection{Comparison with the state of the art}
\label{comparison-with-sota}
Finally, in Table \ref{tab:sota-vad-auc} we compare MOVAD (with $320\times240$ and $640\times480$ input videos size) with state-of-the-art (SOTA) models.
The training of MOVAD followed the best configuration found using information obtained through ablation studies: VCL of 8, 3 LSTM cells and $\mathit{NF}=4$.
Both MOVAD configurations surpass the SOTA models.
Our best MOVAD surpasses SOTA by +$2.87$ AUC, reaching the highest value of $82.17\%$ AUC.
Table \ref{tab:sota-vad-auc-per-class} shows results per class, grouped by ego and non-ego anomaly categories (see \cite{9712446} for explanation of classes).
It is worth to note that, unlike previous SOTA models which incorporate e.g. object detectors, our model also processes without any problem videos in which traffic participants are not present or visible. 
For this reason, the table shows also the ego-involved VO (vehicle-obstacle collision) and OO (oncoming out-of-control) classes.
    \section{Conclusions}
\label{sec:conclusions}

In this paper, we propose MOVAD, a new architecture for the frame-level VAD task, capable to work in an online fashion, handling the most restrictive VAD scenario with an end-to-end training, requiring only RGB frames.
It is composed by STMM, which extracts information related to the ongoing action, implemented by a VST, and LTMM that considers remote past, thanks to LSTM injected inside the classifier.
We evaluated its performance on DoTA dataset, a collection of dash-mounted camera videos of accidents, reaching $82.17\%$ AUC, surpassing SOTA by +$2.87$ AUC.
    \bibliographystyle{splncs04}
    \bibliography{main}
\end{document}